\documentclass[conference]{IEEEtran}
\IEEEoverridecommandlockouts
\usepackage{cite}
\usepackage{amsmath,amssymb,amsfonts}
\usepackage{algorithmic}
\usepackage{graphicx}
\usepackage{textcomp}
\usepackage{xcolor}

\usepackage{multirow}

\def\BibTeX{{\rm B\kern-.05em{\sc i\kern-.025em b}\kern-.08em
    T\kern-.1667em\lower.7ex\hbox{E}\kern-.125emX}}
\begin{document}

\title{Dual Stream Independence Decoupling for True Emotion Recognition under Masked Expressions}

\author{
	\IEEEauthorblockN{Jinsheng Wei, Xiguang Zhang, Zheng Shi \textsuperscript{*},  Guanming Lu}
	\IEEEauthorblockA{Nanjing University of Posts and Telecommunications, Nanjing, China}
	\thanks{*Corresponding author}
}

\maketitle

\begin{abstract}
Recongnizing true emotions from masked expressions is extremely challenging due to deliberate concealment. Existing paradigms recognize true emotions from masked-expression clips that contain onsetframes just starting to disguise. However, this paradigm may not reflect the actual disguised state, as the onsetframe leaks the true emotional information without reaching a stable disguise state. Thus, this paper introduces a novel apexframe-based paradigm that classifies true emotions from the apexframe with a stable disguised state. Furthermore, this paper proposes a novel dual stream independence decoupling framework that decouples true and disguised emotion features, avoiding the interference of disguised emotions on true emotions. For efficient decoupling, we design a decoupling loss group, comprising two classification losses that learn true emotion and disguised expression features, respectively, and a Hilbert-Schmidt Independence loss that enhances the independence of two features. Experiments demonstrate that the apexframe-based paradigm is challenging, and the proposed decouple framework improves recogntioni performances.


\end{abstract}


\section{Introduction}
\label{sec:intro}

Facial expressions reflect human emotional states \cite{zhao2023facial,wei2025multi}. In social interactions, people may conceal their true emotions through the masked expressions of deliberate disguise. Here, masked expressions refer to the facial appearance when disguised emotions are inconsistent with their true emotions \cite{Kulkarni2021Automatic,yan2014casme,cai2024mfdan,nguyen2023micron}, while masked emotions are the emotions that individuals deliberately want to express. Researches \cite{ekman2003unmasking,mo2021mfed} demonstrate that there are differences between masked expressions and natural expressions, and these differences reflect their true emotions. Therefore, Recognizing True Emotions (TER) underlying masked expressions is feasible and has critical applications in security, healthcare, and human-computer interaction \cite{zhang2023recognition,li2022deep,wei2025multi}.  Due to deliberate concealment, masked expressions contain rich disguised emotional information, while containing less true emotional information. Therefore, TER is extremely challenging.

\begin{figure}[t]
\centering{\includegraphics[scale=0.25]{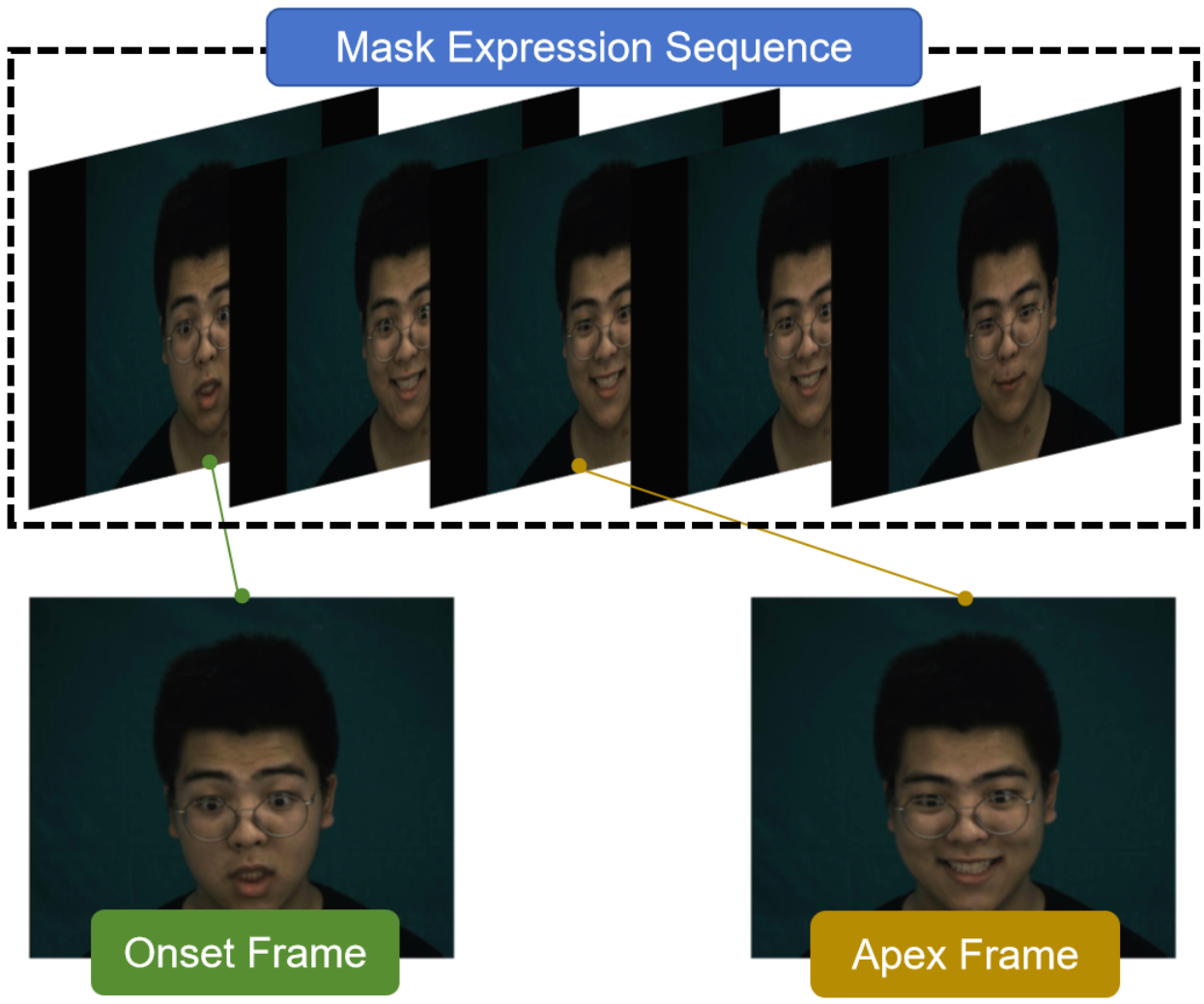}}
\caption{The illustration of onset and apex frames from the masked expression sequence: a "happiness" disguised emotion with a "surprise" true emotion. The onset frame reveals the true emotion (surprise), while the apex frame reveals the disguised emotion (happiness).  }
\label{fig-teaser}
\end{figure}

To support research in this area, Mo et al. \cite{mo2021mfed} constructed the Masked Facial Expression Database (MFED) dataset. In MFED, the subjects first watch emotional stimulation videos to induce real emotions (called experienced emotion in MFED), and then display specific disguised expressions (called required expression in MFED) based on the prompt images. When the two are not consistent, a masked expression is formed. Each recorded sequence is annotated with an onset frame (beginning of masked expression), an apex frame (the peak intensity of masked expression), and an offset frame (masked expression end). As shown in Fig. \ref{fig-teaser}, the onset frame contains rich true emotional information, while the apex frame contains rich disguised emotional information \cite{liu2023recognition}. Thus, existing methods achieved satisfactory performance in learning true emotional features from the onset frame or video clips containing the onset frame.
However, in the onset frame, the masked expression is just beginning to be revealed and does not fully conceal the true emotions. Thus, the onset frame-based TER paradigm suffers from a fundamental limitation: the onset frame captures the subject's true emotional response before the disguised emotion stabilizes, thus leaking true emotion information and simplifying the TER task. As shown in Fig. \ref{fig-teaser}, the onset frame mainly expresses the true emotion ("surprise"). In the apex frame, the masked expression reaches its peak and stable state, and the apex frame-based TER presents a more realistic and challenging setting for true emotion inference from masked expressions. For example, the apex frame in Fig. \ref{fig-teaser} expresses rich disguised emotion ("Happiness). So far, due to the dominance of disguised emotions in apex frames, the contribution of the apex frame for true emotional analysis is not clear. Therefore, this paper explores the contribution of apex frames to TER and proposes a new paradigm that aims to classify true emotions from the apex frame. In this paradigm, the disguised emotions have been fully expressed, which is more consistent with practical scenarios.

In a disguised state, disguised emotions dominate masked expressions to conceal true emotions \cite{he2025weighted, wei2022learning,yang2025micro}. Disguised emotion information seriously interferes with the model's ability to mine true emotional information. Accurately recognizing true emotions requires separating overlapping emotional signals present in masked expressions. The existing methods \cite{zhang2025masked,he2025weighted,mo2021mfed} design the classification model to violently extract true emotion information using category labels, but cannot effectively eliminate the interference of disguised emotion information on true emotional information. Thus, this paper designs a Dual Stream Independence Decoupling (DSID) model that effectively decouples disguised emotion features and true emotion features from masked expressions. 

After extracting global emotion features, the model introduces two branches to learn true emotion features and disguised emotion features, respectively.
To achieve decoupling, a decoupling loss group is designed and includes two classification losses and one independence loss. Specifically, for the two branches, two emotion classification losses are designed to learn their unique features. Furthermore, to improve the decoupling performance, inspired by the Hilbert Schmidt Independence Criterion (HSIC) \cite{ma2020hsic,gretton2005measuring,albert2022adaptive}, a Hilbert Schmidt Independence Criterion loss is designed to enhance the independence of the true and disguised emotion features.

Overall, the contributions of this paper are as follows:
\begin{itemize}
  \item A new paradigm is introduced to shift the focus from onset-frame to apex-frame analysis, establishing a more realistic and challenging benchmark for true emotion recognition under masked expressions.
  \item A novel decoupling framework is proposed to isolate true and disguised emotion features, reducing interference between them.
  \item An hilbert schmidt independence-based decoupling loss group is designed to enhance the feature independence between the true and disguised emotions, supported by dedicated classification losses for each branch.
\end{itemize}
%
\begin{figure*}[t]
\centering{\includegraphics[scale=0.4]{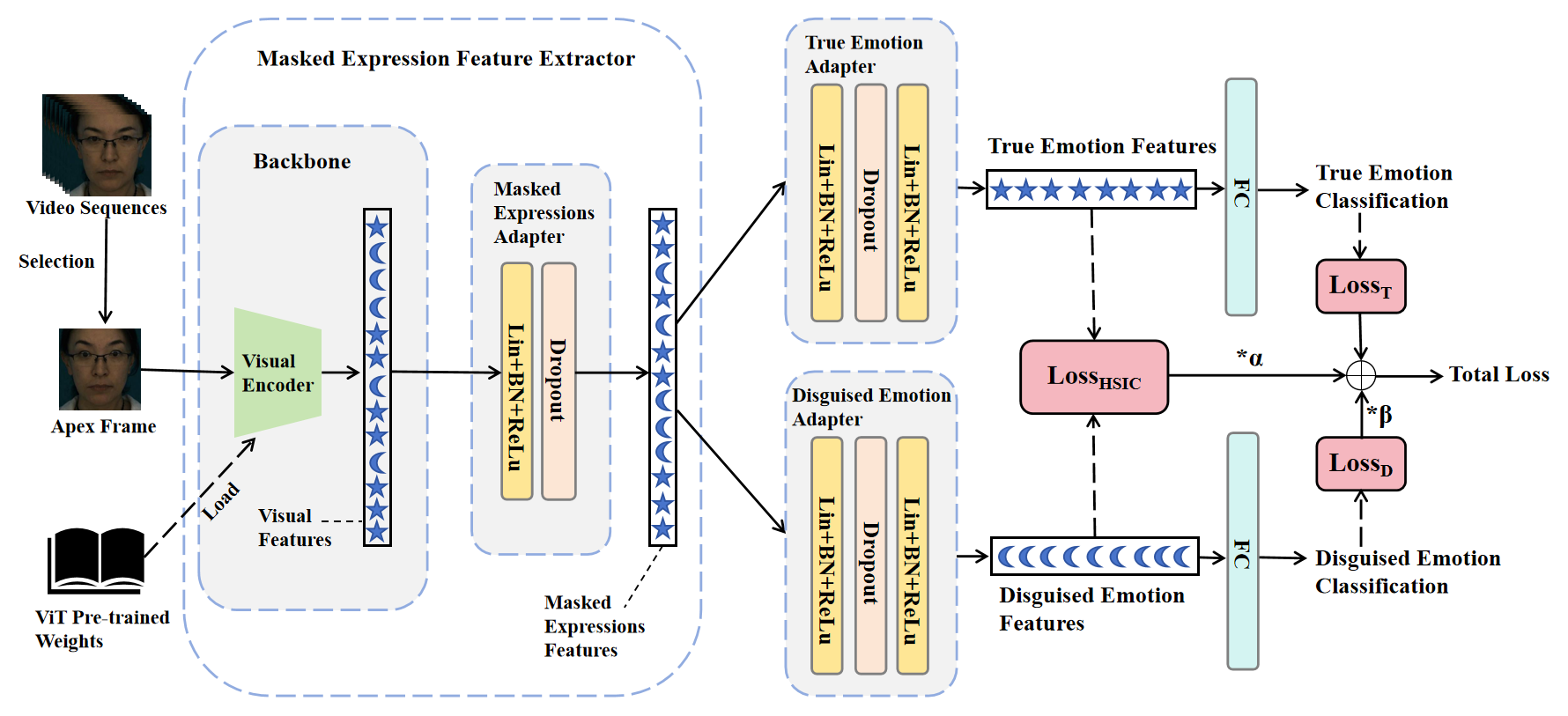}}
\caption{Framework of dual stream independence decoupling model.}
\label{fig-framework}
\end{figure*}
\section{Related Works}

\subsection{Recognition Paradigm}

The appropriate selection of input frames is a crucial factor affecting the performance of masked expression recognition. 
Zhou et al. \cite{zhou2024seeing} used a time-interpolation model to unify the input into a sequence of 30 frames in their study, which, while preserving complete temporal dynamic information, also introduced computational burden and redundant information. To further improve efficiency and mitigate data redundancy, Gu et al. \cite{gu2024kf} proposed an input strategy based on keyframe selection, using temporal binning and Euclidean distance calculation to select the most representative keyframes from each bin to form a compact sequence, reducing computational load while maintaining the integrity of temporal information. Liu et al. \cite{liu2023recognition} explicitly used the onset frame for true emotion recognition tasks. They believe that the onset frame retains more true emotional information, while the apex frame was used for disguised emotion recognition. 

Although the above methods improve the TER performance to a certain extent, their adopted paradigm faces the true emotion leakage issue. Different from these works, this study explores a novel apexframe-based paradigm that captures true emotion cues from the apex frames of masked expression videos, and in line with more challenging real-world scenarios.

\subsection{Masked Expression Recognition}
Masked Expression Recognition can be divided into two parts, namely, TER and Disguised Emotion Recognition (DER) based on Masked Expressions.
 Liu et al. \cite{liu2023recognition} proposed a masked expression recognition method based on transfer learning, using a ResNet18 model pre-trained on the ImageNet dataset as the backbone network. This method redesigned the classifier and introduced a Dropout layer, while also applying various data augmentation techniques to enhance the model's robustness and generalization ability.
Zhang et al. \cite{zhang2025masked} present an end-to-end dual-stream network that joints the Transformer and the graph convolutional network (GCN) \cite{kipf2016semi} to model image features and AU features, respectively, achieving the masked expression recognition.
  He et al. \cite{he2025weighted} proposed a multi-task spatiotemporal weighting network (MTSTWN) that integrates spatiotemporal feature modulation and an adaptive spatial attention module, achieving state-of-the-art performance in masked facial expression recognition.

Different with above works, this works make the first step to decouple the true and disguised emotion features. The proposed decoupling framework with a two stream structure introduces the Hilbert-Schmidt Independence Criterion (HSIC) as a feature decoupling constraint that effectively promotes their evolution towards orthogonal independence in the feature space.

\section{Method}

\subsection{Framework}
The proposed framework consists of a Masked Expression Feature Extractor (MEFE) and dual adaptation branches, designed to decouple features of true and disguised emotions. The workflow is as shown in Fig. \ref{fig-framework}. The details are as follows:
\paragraph{Backbone}
In this work, Vision Transformer (ViT) \cite{dosovitskiy2020image} is employed as the backbone, initialized with the pre-trained weights on the CLIP's visual encoder \cite{radford2021learningCLIP}. We first select the apex frame from the mask expression videos and feed it into the visual encoder. Based on the learned visual representation ability in pre-trained weights, this backbone can capture high-level semantic information of facial expressions and provide more comprehensive visual semantic knowledge for subsequent task adapters.
\paragraph{Masked Expression Adapter}
To adapt the masked expression recognition task, a Masked Expression Adapter is designed to map the visual feature obtained by the backbone to masked expression features. The masked expression adapter comprises Linear, BN, ReLU and Dropout layers serially. The output of this adapter is shared as the input for subsequent dual branches.
\paragraph{Dual Emotion Adapters}
Two parallel adapters, the true emotion adapter and the disguised emotion adapter, are designed to extract true and disguised emotion features, respectively. The true emotion adapter comprises Linear, BN, ReLU and Dropout layers serially to process masked expression features, learning true emotion features for true emotion classification. The disguised emotion adapter uses the same architecture as the true emotion adapter to learn disguised emotion features for disguised emotion classification.

\subsection{Decoupling Loss Group}
In order to enable the proposed framework to have decoupling capability, this paper proposes a decoupling loss group, including true emotion classification loss $\mathcal{L}_{\text{T}}$, disguised emotion classification loss $\mathcal{L}_{\text{D}}$, and HSIC loss ($\mathcal{L}_{\text{HSIC}}$). By jointing the three losses, the total loss are calculated by: 

\begin{equation}
\mathcal{L}_{\text{Total}} = \mathcal{L}_{\text{T}} + \beta \cdot \mathcal{L}_{\text{D}} + \alpha \cdot \mathcal{L}_{\text{HSIC}},
\label{eq}
\end{equation}

where $\alpha$ and $\beta$ are trade-off hyperparameters to balance the discrimination and independence of two emotion features.

\paragraph{Two Emotion Classification Losses ($\mathcal{L}_{\text{T}}$, $\mathcal{L}_{\text{D}}$)}
\begin{itemize}
    \item $\mathcal{L}_{\text{T}}$: Cross-entropy loss is employed to learn the unique features of true emotions and as the true emotion classification loss. This loss is calculated by penalising the distance between the predicted labels of the True Emotion Adapter and the ground-truth labels of true emotions. 
    \item $\mathcal{L}_{\text{D}}$: Consistent with the true emotion classification loss, the disguised emotion classification loss also employs cross-entropy los to learn the unique features of disguised emotions.
    
\end{itemize}
\paragraph{HSIC Feature Independence Loss ($\mathcal{L}_{\text{HSIC}}$)}
To decouple the true emotion features and disguised emotion features extracted by the dual-branch adapters, we designed the HSIC Loss ($\mathcal{L}_{\text{HSIC}}$), which quantifies the statistical independence between two feature distributions \cite{gretton2005measuring}. The core principle of HSIC loss is to minimize the statistical dependence between the two feature. when the true emotion features and disguised emotion features are completely decoupled (statistically independent), the HSIC value converges to 0.

For the $i$-th sample in a batch, let $X_i \in \mathbb{R}^d$ denote the true emotion feature (dimension $d$) and $Y_i \in \mathbb{R}^d$ denote the disguised emotion feature (with the same dimension to ensure kernel compatibility). We first perform $L_2$ normalization on all feature vectors to enhance numerical stability during kernel calculation, yielding normalized features $\hat{X}_i = \frac{X_i}{\|X_i\|_2}$ and $\hat{Y}_i = \frac{Y_i}{\|Y_i\|_2}$.

The HSIC value for the feature pair $(\hat{X}_i, \hat{Y}_i)$ is approximated using a kernel-based method, where two types of kernels are supported (Radial Basis Function, RBF, and Linear kernel):
\begin{equation}
\text{HSIC}(\hat{X}_i, \hat{Y}_i) = \left(k_{\hat{X}_i\hat{Y}_i} - k_{\hat{X}_i\hat{X}_i} \cdot k_{\hat{Y}_i\hat{Y}_i}\right)^2
\end{equation}
where:
\begin{itemize}
    \item $k_{\hat{X}_i\hat{Y}_i}$ represents the cross-kernel value between the normalized true emotion feature $\hat{X}_i$ and normalized disguised emotion feature $\hat{Y}_i$.
    \item $k_{\hat{X}_i\hat{X}_i}$ (self-kernel of $\hat{X}_i$) and $k_{\hat{Y}_i\hat{Y}_i}$ (self-kernel of $\hat{Y}_i$) denote the kernel values of each feature vector with itself.
\end{itemize}
In this paper, we select the RBF kernel, and the kernel value between two vectors $\mathbf{u}$ and $\mathbf{v}$ is calculated as:
          \begin{equation}
          \text k_{\mathbf{uv}}^{\text{RBF}} = \exp\left(-\frac{\|\mathbf{u} - \mathbf{v}\|_2^2}{2\sigma^2}\right),
          \label{rbf-k}
          \end{equation}
          where $\sigma$ is the bandwidth hyperparameter controlling the smoothness of the kernel.

The physical meaning of \eqref{rbf-k} is that when $\hat{X}_i$ and $\hat{Y}_i$ are independent, the cross-kernel $k_{\hat{X}_i\hat{Y}_i}$ approximates the product of the self-kernels $k_{\hat{X}_i\hat{X}_i} \cdot k_{\hat{Y}_i\hat{Y}_i}$, making the HSIC value close to 0. Conversely, if the two features are dependent, the HSIC value increases significantly.

After computing the HSIC value for each sample pair in the batch, these HSIC values are aggregate to obtain the batch-level HSIC loss:
\begin{equation}
\mathcal{L}_{\text{HSIC}} = \frac{1}{N} \sum_{i=1}^N \text{HSIC}(\hat{X}_i, \hat{Y}_i),
\end{equation}
where $N$ is the batch size.

\section{Experiments}

\subsection{Dataset and Preprocessing}
All experiments in this paper are conducted on the Masked Facial Expression Database (MFED) \cite{mo2021mfed} constructed by the Institute of Psychology, Chinese Academy of Sciences. This database contains 778 video sequences contributed by 22 subjects (10 males, 12 females, aged 19–26 years), with a resolution of 1280×720 pixels and a frame rate of 25 fps. During the data acquisition phase, subjects first watched videos designed to induce six basic emotions (anger, disgust, fear, happiness, sadness, surprise), and then immediately made an expression according to on-screen instructions. If this expression did not match the true emotion induced by the video, it was defined as a masked expression, and the complete video sequence from receiving the instruction to completing the expression was recorded.

To focus on the core task of true emotion recognition, we removed sequences where the true emotion and the instructed expression were consistent (i.e., normal expressions), retaining only the data for all 30 types of masked expressions (six true emotions × five corresponding inconsistent disguised expressions). All videos have been annotated with the onset, apex and offset frames of the masked expression dynamics.

In the preprocessing stage, we first performed face detection and background cropping on each video \cite{xie2022overview}. Then, we extracted the apex frame of the standardized sequence as the static image input for the model. In the study of TER paradigm, the onset is extracted as the input of the onset-based paradigm.
All processed images were uniformly scaled to 224×224 pixels.

\subsection{Experimental Setup}
This study employs a Leave-One-Subject-Out (LOSO) cross-validation strategy \cite{gu2024kf,wei2023geometric} to evaluate model performance. Model performance is evaluated using accuracy and F1 score.

The experiments were conducted on a hardware platform equipped with a 12th generation Intel Core i7-12700KF processor and an NVIDIA GeForce RTX 3060Ti GPU. During the data preprocessing stage, data augmentation techniques such as random horizontal flipping, random rotation ($±10^{\circ}$ ), and color jittering were used. Image normalization used the statistics of the ImageNet dataset (mean=[0.485, 0.456, 0.406], standard deviation=[0.229, 0.224, 0.225]).

The max training epoch and batch size are set to 200 and 32, respectively. The dropout probability is 0.5. The Adam optimizer was used with an initial learning rate of 0.0005 and a weight decay (L2 regularization) value of 0.0005. To prevent overfitting, an early stopping strategy was employed; training was terminated when the validation set accuracy did not improve for 50 consecutive epochs.

\subsection{The Study of TER Paradigm}
Tab. \ref{tab:paradigm} reports the performance comparison between the onset frame-based paradigm and the apex frame-based paradigms. Both paradigms use the same model that employs the advanced ViT as the backbone with an independent classification pipeline. Tab. \ref{tab:paradigm} shows that the disguised emotion recognition based on apex frames achieves a significantly better performance than that based on onset frames, while the opposite is true for true emotion recognition.  It indicates that the information contained in the two types of frames has significant differences. 
Specifically, the onset frame contains rich true emotion information, while the apex frame is more dominated by disguised emotion information, which verifies the viewpoint presented in the previous section. 

Furthermore, in terms of true emotion recognition, the apex frame-based paradigm achieved poorer performance than the onset frame-based paradigm. It demonstrates that in the apex frame, the disguised emotion produces a stronger interference on true emotion cues, and the apex frame-based paradigm proposed in this paper is more challenging and has practical application value. At the same time, this also confirms that the onset frame has not yet reached a stable disguised state, and there is a leakage of true emotional information.
\begin{table}[t]
  \centering
  \caption{The comparison results between the apex frame-based paradigm and the onset frame-based paradigm.}
  \label{tab:paradigm}
  \small
  \setlength{\tabcolsep}{4pt}
  \begin{tabular}{|c|c|c|c|c|}
    \hline
    \multirow{2}{*}[0pt]{\textbf{Input}} & \multicolumn{2}{c|}{\textbf{TER}} & \multicolumn{2}{c|}{\textbf{DER}} \\
    \cline{2-5}
    & \textbf{\textit{Accuracy}} & \textbf{\textit{F1-Score}}& \textbf{\textit{Accuracy}} & \textbf{\textit{F1-Score}} \\
    \hline
    Apex-based & 0.3719 & 0.3518 & \textbf{0.4707} & \textbf{0.4604}\\
    \hline
    Onset-based & \textbf{0.4136} & \textbf{0.3933}  & 0.3781& 0.3725\\
    \hline
  \end{tabular}
\end{table}

\subsection{Ablation Study}
This ablation study validates the effectiveness of the proposed decoupling framework (DSID) by comparing it with a standard Vision Transformer (ViT) with no decoupling, and an ablated DSID model trained without the HSIC loss (DSID w/o HSIC).
﻿
As shown in Tab, the results clearly demonstrate the distinct contributions of the decoupling structure and the independence constraint. For DER, DSID w/o HSIC is surpier to ViT model, as seen in the significant jump from ViT (46.04\% F1) to DSID w/o HSIC (51.18\% F1). It demonstrates that the dual stream structure can guide the separation of features through two classification losses, effectively improving the model to capture the disguised emotional clues. However, for the more challenging TER, the architectural separation alone yields only a minor improvement. Furthermore. DSID achieves a better performance compared to DSID w/o HSIC, especially on TER task. The critical gain comes from the HSIC loss, which actively enforces feature independence, pushing TER performance from 36.83\% to 37.07\% F1-score. This shows that structural separation is insufficient; statistical independence is essential to isolate subtle genuine emotions from the overpowering disguise.
\begin{table}[t]
  \centering
  \caption{Ablation Study: Feature Decoupling vs No Decoupling}
  \label{tab:ablation}
  \resizebox{\linewidth}{!}{
  \begin{tabular}{|c|c|c|c|c|}
    \hline
    \multirow{2}{*}[0pt]{\textbf{Methods}} 
      & \multicolumn{2}{c|}{\textbf{TER}} 
      & \multicolumn{2}{c|}{\textbf{DER}} \\
    \cline{2-5}
    & \textbf{\textit{Accuracy}} 
    & \textbf{\textit{F1-Score}} 
    & \textbf{\textit{Accuracy}} 
    & \textbf{\textit{F1-Score}} \\
    \hline
    ViT   & 0.3719 & 0.3518 & 0.4707 & 0.4604 \\
    \hline
    DSID w/o HSIC   & 0.3781 & 0.3683 & 0.5154 & 0.5118 \\
    \hline
    DSID & \textbf{0.3920} & \textbf{0.3707} & \textbf{0.5170} & \textbf{0.5166} \\
    \hline
  \end{tabular}
  }
\end{table}

\begin{table}[t]
  \centering
  \caption{Prameter evaluation on $\alpha$.}
  \label{tab:alpha_tune}
  \begin{tabular}{|c|c|c|c|c|}
    \hline
    \multirow{2}{*}[0pt]{\textbf{$\alpha$}} 
      & \multicolumn{2}{c|}{\textbf{TER}} 
      & \multicolumn{2}{c|}{\textbf{DER}} \\
    \cline{2-5}
    & \textbf{\textit{Accuracy}} 
    & \textbf{\textit{F1-Score}} 
    & \textbf{\textit{Accuracy}} 
    & \textbf{\textit{F1-Score}} \\
    \hline
    0.1 & 0.3796 & 0.3677 & \textbf{0.5432} & \textbf{0.5389} \\
    \hline
    0.3 & 0.3873 & 0.3703 & 0.5077 & 0.5073 \\
    \hline
    0.5 & \textbf{0.3920} & \textbf{0.3707} & 0.5170 & 0.5166 \\
    \hline
    0.7 & 0.3796 & 0.3665 & 0.5062 & 0.5077 \\
    \hline
    0.9 & 0.3688 & 0.3446 & 0.4985 & 0.4991 \\
    \hline
    1.0 & 0.3673 & 0.3419 & 0.5062 & 0.5038 \\
    \hline
  \end{tabular}
\end{table}

\begin{table}[t]
  \centering
  \caption{Prameter evaluation on  $\beta$.}
  \label{tab:beta_tune}
  \begin{tabular}{|c|c|c|c|c|}
    \hline
    \multirow{2}{*}[0pt]{\textbf{$\beta$}} 
      & \multicolumn{2}{c|}{\textbf{TER}} 
      & \multicolumn{2}{c|}{\textbf{DER}} \\
    \cline{2-5}
    & \textbf{\textit{Accuracy}} 
    & \textbf{\textit{F1-Score}} 
    & \textbf{\textit{Accuracy}} 
    & \textbf{\textit{F1-Score}} \\
    \hline
    0.1 & 0.3657 & 0.3580 & 0.3596 & 0.3592 \\
    \hline
    0.3 & 0.3843 & \textbf{0.3750} & 0.4907 & 0.4908 \\
    \hline
    0.5 & 0.3765 & 0.3583 & 0.4861 & 0.4809 \\
    \hline
    0.7 & 0.3827 & 0.3631 & 0.4954 & 0.4939 \\
    \hline
    0.9 & 0.3750 & 0.3586 & 0.5139 & 0.5137 \\
    \hline
    1.0 & \textbf{0.3920} & 0.3707 & \textbf{0.5170} & \textbf{0.5166} \\
    \hline
  \end{tabular}
\end{table}

%

\begin{table}[t]
  \centering
  \caption{Comparing with other models}
  \label{tab:backbone_compare}
  \begin{tabular}{|c|c|c|}
    \hline
    \multirow{2}{*}[0pt]{\textbf{Methods}} 
      & \multicolumn{2}{c|}{\textbf{True Emotion}} \\
    \cline{2-3}
    & \textbf{\textit{Accuracy}} 
    & \textbf{\textit{F1-Score}} \\
    \hline
    ResNet18 \cite{he2016deep} & 0.2948 & 0.2869 \\
    \hline
    ViT \cite{dosovitskiy2020image}& 0.3719 & 0.3518 \\
    \hline
    CLIP \cite{radford2021learningCLIP}& 0.3441 & 0.3319 \\
    \hline
    DSID & \textbf{0.3920} & \textbf{0.3707} \\
    \hline
  \end{tabular}
\end{table}

\subsection{Parameter Evaluation}
\paragraph{The Evaluation on \(\alpha\)}
As shown in Tab. \ref{tab:alpha_tune}, the HSIC loss weight $\alpha$s is evaluated ranging from 0.1 to 1 with a 0.2 interval, while keeping the disguised emotion classification loss weight $\beta$ fixed at 1.

For the TER task, as $\alpha$ increases from 0.1 to 0.5, accuracy improves from 0.3796 to a peak of 0.3920, and the F1-score rises from 0.3677 to 0.3707. This demonstrates that a moderate independence constraint can guide the model to separate true emotional features from the disguised expression. However, when $\alpha$ exceeds 0.5, performance declines, with notable degradation at $\alpha$=0.9 and 1.0 (accuracy drops to 0.3688 and 0.3673, respectively). This suggests that an excessively strong independence constraint may disrupt discriminative feature information, impairing TER.

For the DER task, the performance is more sensitive to $\alpha$. The optimal $\alpha$ is 1 with a highist performance (accuracy 0.5432, F1-score 0.5389). As $\alpha$ increases, performance generally decreases, reaching its lowest point at $\alpha$=0.9 (accuracy 0.4985). It turns out that while the HSIC loss enhances feature independence, it may also suppress the learning of disguised emotion features, especially when weighted too heavily. 

Overall, $\alpha$=0.5 emerges as an optimal balance for both tasks. When $\alpha$ is 0.5, TER performance peaks, while DER remains relatively strong (accuracy 0.5170), albeit slightly below its maximum at $\alpha$=0.1 (a difference of 2.6 percentage points). This configuration achieves an effective trade-off, namely, sufficient feature decoupling to improve TER, without critically compromising DER discriminability.

\paragraph{The Evaluation on \(\beta\)}
This experiment evaluates the impact of the disguised emotion classification loss weight \(\beta\) on model performance, and the $\alpha$ is fixed at 0.5 (optimal value).

Overall, TER is sensitive to $\beta$, but there is no clear regularity. Furthermore, when $\beta$ is set to 0.1, the performances of both TER and DER have an obvious decrease, especially on DER. It demonstrates that a too small  $\beta$ makes it difficult for the model to learn stable and discriminative disguised emotional features, thereby leading to interference in both tasks under the constraint of independence loss. Also,  Performance on DER shows a steady improvement as $\beta$ increases from 0.1 to 1.0, which demonstrates that the strong supervision based on $\beta$ is positively correlated with the mining of disguised emotions.

\subsection{Comparing With Other Models}

Tab. \ref{tab:backbone_compare} reports the comparison results between the proposed method and other models. All results were obtained under the setting of apexframe-based TER frames. Since there is currently no work based on this paradigm, several representative models were selected as comparison methods. 

Overall, the proposed DSID framework achieves the best results, outperforming ResNet18, ViT, and CLIP. Specifically, traditional ResNet18, as a representative of convolutional neural networks, has the lowest performance (accuracy of 0.2948), indicating that relying solely on local convolutional features is difficult to capture subtle true emotional clues from highly disguised apex frames. The Vision Transformer (ViT) has significantly improved its performance with its global self-attention mechanism. It turns out that modeling long-range dependencies is crucial for understanding complex facial expressions. Furthermore, the multimodal pre-trained model CLIP performs better than ResNet18, but not as well as ViT. It demonstrates that CLIP's visual encoder has limited discriminative power for true emotion features without specific task tuning. Also, the fine-grained level of text may also affect its ability to capture small emotion differences.

DSID holds the advantage of its targeted feature decoupling and independence constraint design and achieves a competitive performance in the apex frame-based TER task, which provides an effective baseline solution for subsequent research.
\subsection{Conclusions}

 This paper presents a novel apex-frame-based paradigm for true emotion recognition from masked expressions, which is a more realistic and challenging task compared to the conventional onset-frame analysis. To tackle this, we propose a Dual-Stream Independence Decoupling (DSID) framework. Its core innovation lies in explicitly separating true emotional features from disguised emotion features via two dedicated branches, enforced by a Hilbert-Schmidt Independence Criterion (HSIC) loss to ensure their statistical independence. The ablation study confirmed the necessity of both the dual-stream structure (crucial for DER) and the HSIC loss (critical for TER). Hyperparameter tuning identified $\alpha$=0.5 and $\beta$=1.0 as the optimal weights, balancing the independence constraint with task-specific supervision. Finally, comparisons with other models demonstrated the superior effectiveness of the well-designed DSID framework for TER task. Collectively, the experiments validate that our method successfully decouples and recognizes true emotions from highly disguised apex frames, establishing a strong baseline for this new research direction.

\bibliographystyle{IEEEbib}
\bibliography{icme2026references}


\end{document}